\documentclass{article}

     \usepackage[preprint, nonatbib]{neurips_2021}

\usepackage[utf8]{inputenc} 
\usepackage[T1]{fontenc}    
\usepackage{url}            
\usepackage{booktabs}       
\usepackage{amsfonts}       
\usepackage{nicefrac}       
\usepackage{microtype}      
\usepackage{xcolor}         
\usepackage{amsmath}
\usepackage{graphicx}
\usepackage{float}
\usepackage{tcolorbox}
\usepackage[style=numeric, sorting=none]{biblatex}
\usepackage{amsthm}
\usepackage{algorithm}
\usepackage{wrapfig}
\usepackage[noend]{algpseudocode}

\usepackage{caption}
\usepackage{subcaption}

\title{Solving hybrid machine learning tasks by traversing weight space geodesics}

\author{
  Guruprasad Raghavan\\
  Caltech\\
  Pasadena, CA 91106 \\
  \texttt{graghava@caltech.edu} \\
  \And
  Matt Thomson\\
  Caltech\\
  Pasadena, CA 91106 \\
  \texttt{mthomson@caltech.edu} \\
}

\begin{document}

\maketitle

\begin{abstract}

Machine learning problems have an intrinsic geometric structure as central objects including a neural network's weight space and the loss function associated with a particular task can be viewed as encoding the intrinsic geometry of a given machine learning problem. Therefore,  geometric concepts can be applied to analyze and understand theoretical properties of machine     learning strategies as well as to develop new algorithms.  In this paper, we address three seemingly unrelated open questions in machine learning by viewing them through a unified framework grounded in differential geometry.  Specifically, we view the weight space of a neural network as a manifold endowed with a Riemannian metric that encodes performance on specific tasks. By defining a metric, we can construct geodesic, minimum length
, paths in weight space that represent sets of networks of equivalent or near equivalent functional performance on a specific task.  We, then, traverse geodesic paths while identifying networks that satisfy a second objective. Inspired by the geometric insight, we apply our geodesic framework to 3 major applications: (i) Network sparsification (ii)  Mitigating catastrophic forgetting by constructing networks with high performance on a series of objectives  and (iii) Finding high-accuracy paths connecting distinct local optima of deep networks in the non-convex loss landscape. Our results are obtained on a wide range of network architectures (MLP, VGG11/16) trained on MNIST, CIFAR-10/100. Broadly, we introduce a geometric framework that unifies a range of machine learning objectives and that can be applied to multiple classes of neural network architectures.

\end{abstract}

%








\section{Introduction}

The geometry of weight manifolds and functional spaces represented by artificial neural networks is an important window to ‘understanding’ machine learning. Many open questions in machine learning, when viewed through the lens of geometry, can be related to finding points or paths of equivalent function in the weight and functional spaces \cite{cohen2020separability, jia2019geometric, hauser2018principles, anderson2017high}. Although geometric structure plays a key role in determining the properties of neural network training, application of methods from differential geometry to machine learning is complicated by the dependence on millions of network parameters, the non-convex structure of loss functions, and the lack of computationally efficient algorithms that harness the geometric structure to traverse weight or functional space \cite{auer1996exponentially, choromanska2015loss}. 

In this paper, we propose a mathematical framework grounded in differential geometry for constructing path-connected sets of deep neural networks that have equivalent functional performance on a task. The key contribution of our paper is that we view the networks' weights space as a pseudo-Riemannian manifold equipped with a distance metric that represents task performance while simultaneously capturing task-independent network properties, like network sparseness. We formalize the ``search'' for a suitable network (based on the application of interest) as a dynamic movement on the curved pseudo-Riemannian manifold \cite{amari1982differential}. Further, we demonstrate that geodesics, minimum length paths, on the network weights space provide high performance paths that the network can traverse to maintain performance while `searching-out' for other networks that satisfy additional objectives. Specifically we develop a procedure based on the geodesic equation to find sets of path connected networks that achieve high performance while also satisfying a second objective like sparsification or mitigating catastrophic interference.

We demonstrate that our framework can be applied to solve three (seemingly unrelated) major problems in machine learning: (i) Discovering sparse counterparts of dense neural networks and high-accuracy paths that connect the two (dense and sparse networks), (ii) Enabling continual learning by mitigating catastrophic forgetting and (iii) Finding high-accuracy paths connecting two trained deep networks (mode-connectivity) in a non-convex loss landscape. These applications, when viewed through the lens of differential geometry can be solved by finding points or paths of equivalent function in the functional space of deep networks.  Broadly, our paper demonstrates that differential geometry can provide a mathematical framework and novel algorithms to unify open problems in machine learning in a common geometric language.

\section{Related work}

The parameters of a neural network encodes the function that maps a set of inputs to outputs. Although the function mapping input/output is crucial for a large number of ML applications, the intractability of the function-space has veered researchers away and instead focus on techniques and analyses that concern the parameters of the neural network. The introduction of information geometry by Amari \cite{amari1982differential} pioneered the analysis of neural networks from the lens of their function and output spaces. Natural gradient descent (NGD) \cite{amari1998natural} utilized the KL distance between functions to efficiently train neural networks, by evaluating the Fisher-information matrix for scaling gradient updates based on parameters informativeness. Since then, many distance metrics\cite{neyshabur2015path}, like Deep relative trust \cite{bernstein2020distance}, the L$^2$ distance between neural network functions in Hilbert space \cite{benjamin2018measuring} have been developed for computing the functional distance between neural networks. In addition, distance metrics and algorithms have been developed to navigate the object manifolds in order to learn the best transformations for pattern recognition applications \cite{simard1998transformation}.

\section{Geodesics: Mathematical Framework}







We develop a mathematical framework grounded in differential geometry for navigating the space of neural networks to discover novel networks that have high task-performance while satisfying additional constraints on task-independent network properties, like sparseness. 


\textbf{Preliminaries}:
We represent a feed-forward neural network (NN) as a smooth,  $\mathbb{C}^\infty$function $f(\mathbf{x};\mathbf{w})$, that maps an input vector, $\mathbf{x} \in \mathbb{R}^{\text{k}}$, to an output vector, $f(\mathbf{x};\mathbf{w})  = \mathbf{y}\in \mathbb{R}^{\text{m}}$. The function, $f(\mathbf{x};\mathbf{w})$, is parameterized by a vector of weights, $\mathbf{w} \in \mathbb{R}^\text{n}$, that are typically set in training to solve a specific task. We refer to $W= \mathbb{R}^n$ as the \textit{weight space} ($W$) of the network, and we refer to $\mathcal{Y} = \mathbb{R}^m$  as the \textit{functional space} \cite{mache2006trends}. We also define a loss function, $L: \mathbb{R}^\text{m} \times\mathbb{R} \rightarrow \mathbb{R}$, that provides a scalar measure of network performance for a given task (Figure 1). Please note that the functional space $\mathcal{Y}$ and the loss space $L$ are \textbf{task-dependent} spaces, while the weights space $W$ is \textbf{task-independent} and encodes network properties, like fraction of non-zero weights. 

We construct a metric tensor ($\mathbf{g}$) to evaluate how infinitesimal movements in the weights space $W$ impacts movement in the functional space ($\mathcal{Y}$), effectively measuring the functional-similarity of networks before and after weight perturbation.  The metric tensor can be applied at any point in $W$ to measure the functional impact of an arbitrary network weights perturbation. 

To construct a metric mathematically, we fix the input, $\mathbf{x}$, into a network and ask how the output of the network, $f(\mathbf{x},\mathbf{w})$, moves on the functional space, $\mathcal{Y}$, given an infinitesimal weight perturbation, $\mathbf{du}$, in $W$ where $\mathbf{w_p} = \mathbf{w_t} + \mathbf{du}$. For an infinitesimal perturbation $\mathbf{du}$, 
\begin{align}
f(\mathbf{x},\mathbf{w_t} +\mathbf{du}) \approx f(\mathbf{x},\mathbf{w_t}) +\mathbf{ J_{w_t}} \ \mathbf{du},
\vspace{-2mm}
\end{align}

where $\mathbf{ J_{w_t}}$ is the Jacobian of $f(\mathbf{x},\mathbf{w_t})$ for a fixed $\mathbf{x}$, $J_{i,j} = \frac{\partial f_i}{\partial w^j}$, evaluated at $\mathbf{w_t}$. We measure the change in functional performance given weight perturbation $\mathbf{du}$ as:
\begin{align}
\label{local metric}
d(\mathbf{w_t},\mathbf{w_p}) = |f(\mathbf{x},\mathbf{w_t})-f(\mathbf{x},\mathbf{w_p})|^2  = \mathbf{du}^T \ (\mathbf{J_{w_t}(\mathbf{x})}^T \ \mathbf{J_{w_t}(\mathbf{x})}) \ \mathbf{du}  = 
 \mathbf{du}^T \ \mathbf{g_{w_t}(\mathbf{x})} \ \mathbf{du}
\end{align}
where $\mathbf{g_{w_t}(\mathbf{x})}$ = $\mathbf{J_{w_t}}$($\mathbf{x})^T$ $\mathbf{J_{w_t}(\mathbf{x})}$ is the metric tensor evaluated at the point $\mathbf{w_t} \in W$ for a single datapoint ($\mathbf{x}$). The metric tensor is an $n \times n$ symmetric matrix that defines an inner product and local distance metric, $\langle \mathbf{du}, \mathbf{du} \rangle_{\mathbf{w}}  = \mathbf{du^T} \ \mathbf{g_{w}(\mathbf{x})} \ \mathbf{du}$, on the tangent space of the manifold, $T_w(W)$ at each $\mathbf{w} \in W$.  Explicitly, 
\begin{equation}
    g_{ij}(\mathbf{x}) = \sum_{k=1}^{m}\frac{\partial f_k(\textbf{x},\mathbf{w})}{\partial \mathbf{w^i}}\frac{\partial f_k(\textbf{x},\mathbf{w})}{\partial \mathbf{w^j}},
\end{equation}where the partial derivatives $\frac{\partial f_k(\textbf{x},\mathbf{w})}{\partial \mathbf{w^i}}$ measure change in functional output of a network given a change in weight. In the appendix, we extend the metric formulation to cases where we  consider a set of N training data points, $\mathbf{X}$, and view $\mathbf{g}$ as the average of metrics derived from individual training examples. $ \mathbf{g_w} = \mathbf{g_w(X)} = \Sigma_{i=1}^N\mathbf{g_w(x_i)}/N$. The metric, $\mathbf{g}$, provides a local measure of \textit{functional distance} on the pseudo-Riemmanian manifold $(W,\mathbf{g})$. At each point in weight space, the metric defines the length, $\langle \mathbf{du}, \mathbf{du} \rangle_{\mathbf{w}}$, of a local perturbation by its impact on the functional output of the network (Figure 1B). 

\begin{figure}[t]  
    \centering
  \centerline{  \includegraphics[width=0.8\linewidth]{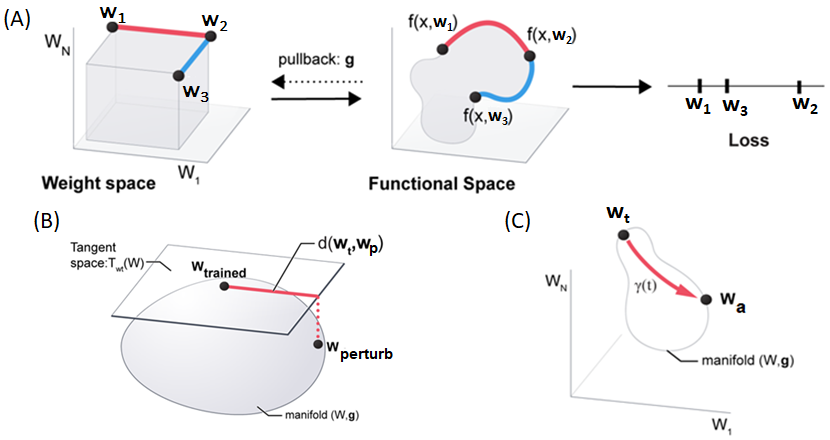}}
    \caption{\textbf{Geometric framework for constructing paths in functional space} (A) Three networks ($\mathbf{w_1},\mathbf{w_2},\mathbf{w_3}$) in weights space $W$ and their relative distance in functional space and loss space. High-performance path is determined by asking how movement in weight space changes functional performance and loss through introduction of a pullback metric $\mathbf{g}$. (B) The metric tensor ($\mathbf{g}$) is evaluated by analyzing the effect of infinitesimal perturbation in the tangent space of a network.(C) Paths between pairs of networks are modeled as long range movement of network weights along a path, $\gamma(t)$, in weight space. }
    \label{fig:schematics}
    \vspace{-4mm}
\end{figure}

Our framework discovers a network that satisfies multiple objectives by constructing a geodesic between two locations in the networks weights space ($W$): one defined by a network that maximizes task performance, $\mathbf{w_t}$, and another defined by a network that satisfies task-independent constraints, the secondary goal, $\mathbf{w_a}$. $\mathbf{w_a}$ could be a single network (if known) or a subspace of networks that satisfy the constraints (if the network is yet to be discovered). 



The global path is constructed in order to simultaneously minimize the movement of the network on the functional space while moving in weights space towards a second point to achieve a secondary goal. We use the metric tensor to determine the functional change across a path-connected set of networks in the networks weights space leading to the second point $\mathbf{w_a}$. Mathematically, the metric changes as we move in $W$ due to the curvature of the ambient space that reflects changes in the vulnerability of a network to weight perturbation (Figure 1C). 

As a network moves along a path $\gamma(t) \in W$ from start network $\mathbf{w_t}$ to the second point encoding the secondary goal $\mathbf{w_a}$, we can analyze the integrated impact on the network performance by using the metric ($\mathbf{g}$) to calculate the length of the path $\gamma(t)$ as: 

\begin{equation}
    S(\gamma) = \int^1 _0 {\langle \frac{d\gamma(t)}{dt}, \frac{d\gamma(t)}{dt}\rangle_{\gamma(t)} } \ dt ,
    \label{eq:geodesicLength}
\end{equation}
where $\langle \frac{d\gamma(t)}{dt}, \frac{d\gamma(t)}{dt}\rangle_{\gamma(t)} = \frac{d\gamma(t)}{dt}^T \mathbf{g}_{\gamma(t)} \frac{d\gamma(t)}{dt}$ is the infinitesimal functional change accrued while traversing path $\gamma(t) \in W$. As the shortest path in functional space - min($S(\gamma$)) is desirable to ensure that the path connected networks are functionally similar, we evaluate the geodesic from $\mathbf{w_t}$ to $\mathbf{w_a}$.


Minimizing $S(\gamma)$ is equivalent to solving the geodesics on $W$ equipped with metric tensor $\mathbf{g}$. 
\begin{align}
    \frac{d^2 w^\eta}{dt^2} + \Gamma_{\mu\nu}^\eta\frac{dw^\mu}{dt} \frac{dw^\nu}{dt} = 0
    \label{eq:geodesicMotion}
\end{align}
where, $w^j$ defines the j'th basis vector of the weights space $W$, $\Gamma_{\mu\nu}^\eta$ specifies the Christoffel symbols $(  \Gamma_{\mu \nu}^\eta =  \sum_r \frac{1}{2}g_{\eta r}^{-1}(\frac{\partial g_{r \mu}}{\partial x^\nu} + \frac{\partial g_{r \nu}}{\partial x^\mu} - \frac{\partial g_{\mu \nu}}{\partial x^r}))$ on the manifold. The Christoffel symbols record infinitesimal changes in the metric tensor ($\mathbf{g}$) along a set of directions on the manifold (Appendix). Since the computation and memory for evaluating Christoffel symbols scales as third order polynomial of network parameters ($\mathcal{O}(n^3)$), we propose an optimization algorithm for evaluating `approximate' geodesics in the manifold.

\textbf{Optimization procedure for approximating geodesics :}

Inspired by the local formulation of the geodesics equation, we introduce an optimization procedure:  Geo($\mathbf{w_t}$, $\mathbf{g}$, $\mathbf{w_a}$, $s$) to construct a global path to find networks that have high performance as well as satisfy additional task-independent constraints. The inputs to the procedure Geo($\mathbf{w_t}$, $\mathbf{g}$, $\mathbf{w_a}$, $s$) are: (i) Start network that maximizes task performance ($\mathbf{w_t}$), (ii) Metric to measure change in task-performance when network moves on weights space ($\mathbf{g}$), (iii) Second network that encodes the secondary goal ($\mathbf{w_a}$) and (iv) User-defined number of steps taken along the path (s). The output of the optimization procedure is the path $\gamma(t)$ beginning from $\gamma(0)$ = $\mathbf{w_t}$ and ending at $\gamma(1) = \mathbf{w_c}$. $\mathbf{w_c}$ could be $\mathbf{w_a}$ for a complete traversal, or could be a different network if the stopping criterion terminates the optimization before the complete traversal.

Starting at $\mathbf{w_t}$, we iteratively solve for $\theta(\mathbf{w})$ using equation-6,7 to traverse the path from $\mathbf{w_t}$ to $\mathbf{w_a}$ in the networks weights space ($W$). $\theta(\mathbf{w})$ is a vector at point $\mathbf{w} \in W$ whose length measures a linear estimate of the change in performance of the network incurred by moving an infinitesimal distance in a given direction in weight space. The procedure finds a direction that (i) minimizes the functional change between networks on a task before and after an update $\{$min: $\langle \theta(\mathbf{w}), \theta(\mathbf{w})\rangle_\mathbf{w}$ = $\theta(\mathbf{w})^T\mathbf{g_w}\theta(\mathbf{w})$ $\}$ while (ii) moving towards the target network ($\mathbf{w_a}$), achieved by maximizing the dot-product of the tangent vector and vector pointing towards $\mathbf{w_a}$: $\{$ max: $\theta(\mathbf{w})^T$ $(\mathbf{w_a}-\mathbf{w})$ $\}$. Having evaluated a suitable $\theta(\mathbf{w})$, we update the networks weights via equation-7, where $\eta$ is the step-size of the update. 

We fix the metric tensor ($\mathbf{g}$) as a representation of the performance on the task as it measures the functional difference before and after every update step, while the direction ($\mathbf{w_a} - \mathbf{w}$) towards $\mathbf{w_a}$ encodes the secondary goal.




\vspace{-3mm}
\begin{equation}
    \text{argmin}_{\theta(\mathbf{w})}  \ \langle \theta(\mathbf{w}), \theta(\mathbf{w}) \rangle_\mathbf{w} - \beta \ \theta(\mathbf{w})^T (\mathbf{w_a}-\mathbf{w}) \\ \text{      subject to:    } \theta(\mathbf{w})^T \theta(\mathbf{w}) \leq 0.01. \\
\end{equation}
\vspace{-4mm}
\begin{align}
    \Delta \mathbf{w} = \eta \frac{\theta(\mathbf{w})}{||\theta(\mathbf{w})||}
\end{align}

The stopping criterion for Geo($\mathbf{w_t}$, $\mathbf{g}$, $\mathbf{w_a}$, $s$) are as follows: (i) Network traverses the entire path from $\mathbf{w_t}$ to $\mathbf{w_a}$, (ii) Number of steps taken reaches user-defined $s$ fed as an input to the procedure and (iii) Network makes small oscillations (moving to and away from $\mathbf{w_a}$). Please note that if stopping criterion (ii) or (iii) is reached, the output of the procedure is the path that terminates at a network different from $\mathbf{w_a}$.



Our optimization procedure is a quadratic program that trades off, through the hyper-parameter $\beta$,  motion towards the target network that encodes the secondary goal ($\mathbf{w_a}$) and the maximization of the functional performance of the intermediate networks along the path (elaborated in the appendix). The strategy discovers multiple paths from the trained network $\mathbf{w_{t}}$ to $\mathbf{w_a}$ (encoding secondary goal) where networks maintain high functional performance during traversal. Of the many paths obtained, we can select the path with the shortest total length (with respect to the metric $\mathbf{g}$) as the best approximation to the geodesic in the manifold. 


\section{Geodesic framework applied to three distinct ML problems }

In the sections that follow, we recast three distinct open questions in ML through the lens of geometry and find solutions by constructing approximate geodesics. The three applications are: (i) Sparsifying networks by traversing geodesics, (ii) Alleviating catastrophic forgetting via geodesics and (iii) Connecting modes of deep neural networks by constructing geodesic paths. 



\subsection{Sparsifying networks by traversing geodesics}

Network sparsification has gained importance in the recent years. Although deep networks are  powerful systems, they require lots of computation and memory making their deployment in resource-constrained environments like mobile phones and smart-devices challenging \cite{blalock2020state}. 

\begin{figure}[t]
    \centering
    \includegraphics[width=1\linewidth]{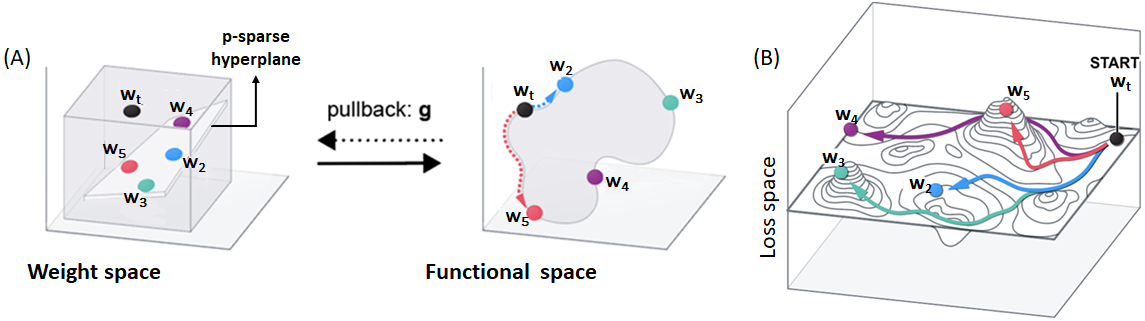}
    \caption{\textbf{Geodesic framework for discovering sparse networks:} (A) Dense network ($\mathbf{w_t}$) and 4 networks ($\mathbf{w_2}, \mathbf{w_3}, \mathbf{w_4}, \mathbf{w_5}$) on the p-sparse hyperplane in the Weight space (left) and their relative distances on the functional manifold (right). (B) $\mathbf{w_t}$, $\mathbf{w_2}$ to $\mathbf{w_5}$ represented on the loss surface. }
    \label{fig:sparseSchematic}
    \vspace{-4mm}
\end{figure}

\begin{wrapfigure}{r}{0.4\textwidth}
    \includegraphics[width=1\linewidth]{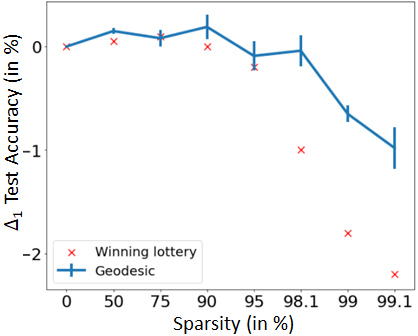}
  \caption{Difference in Top-1 $(\Delta_1)$ test accuracy between the discovered sparse networks and their reference dense network. A comparison with the lottery ticket hypothesis is presented.}
  \label{fig:sparse_results2}
  \vspace{-3mm}
\end{wrapfigure}

Here, we aim to discover a network that simultaneously has (i) high performance on a particular task (eg. CIFAR-10 classification) and (ii) satisfies sparsity constraints, by setting a fraction of its weights to zero. Our optimization procedure (Geo($\mathbf{w_t}$, $\mathbf{g}$, $\mathbf{w_a}$, $s$)) addresses both the objectives by (i) setting the start point $\mathbf{w_t}$ to a dense trained network on the task, followed by computing the metric tensor ($\mathbf{g}$) for the task (eg CIFAR-10 classification)  to evaluate change in task-performance while (ii) moving towards a target network $\mathbf{w_a}$:= p-sparse network, which has p$\%$ of its networks' weights set to zero (encodes the additional sparsity constraint). 



Here, the target sparse network is yet to be discovered, so we designate $\mathbf{w_a}$ as a member of the subspace of networks that satisfies the sparsity constraint. So, we define a p-sparse subspace ($H_p$) in the networks' weights space as a collection of all networks with p$\%$ of their weights set to zero. $H_p = \{ \mathbf{w} \in \mathcal{R}^n : ||\mathbf{w}||_0 = \frac{np}{100} \}$. We choose $\mathbf{w_a}$ to be the projection of the dense network on $H_p$; $\mathbf{w_a}$ = proj($\mathbf{w_t}, H_p$). We constantly update our target network to proj($\mathbf{w}$, $H_p$) every $n_s$ steps taken by the network. 

\begin{algorithm}
\caption{Discovering sparse networks via geodesics}
\begin{algorithmic}[1]
\While{$\mathbf{w_t} \not\in H_p$} \Comment{If start network isn't on p-sparse hyperplane}
    \State $\mathbf{w_a} \leftarrow proj(\mathbf{w_t}, H_p$)  \Comment{Project network on p-sparse hyperplane}
    \State $\gamma(t) \leftarrow Geo(\mathbf{w_t}, \mathbf{g}, \mathbf{w_a}, n_s$) 
    \State $\mathbf{w_c} \leftarrow \gamma(1)$  \Comment{End point of geodesic after taking $n_s$ steps }
    \State $\mathbf{w_t} \leftarrow \mathbf{w_c}$  \Comment{New start network to evaluate geodesic}
\EndWhile
\end{algorithmic}
\end{algorithm}




Figure-\ref{fig:sparseSchematic}A illustrates our adaptation of the geodesic framework for finding functionally similar p-sparse networks by constructing a geodesic to the p-sparse subspace ($H_p$). Here, $\mathbf{w_t}$ is the trained dense network, and $\mathbf{w_2}$, $\mathbf{w_3}$, $\mathbf{w_4}$ and $\mathbf{w_5}$ are p-sparse networks on $H_p$. Their corresponding positions in the functional space highlight the functional closeness of some p-sparse networks to the trained network over others. In figure-\ref{fig:sparseSchematic}B, p-sparse network - $\mathbf{w_2}$ is functionally closest to $\mathbf{w_t}$. 

\begin{figure}[t]
    \centering
    \includegraphics[width=0.9\linewidth]{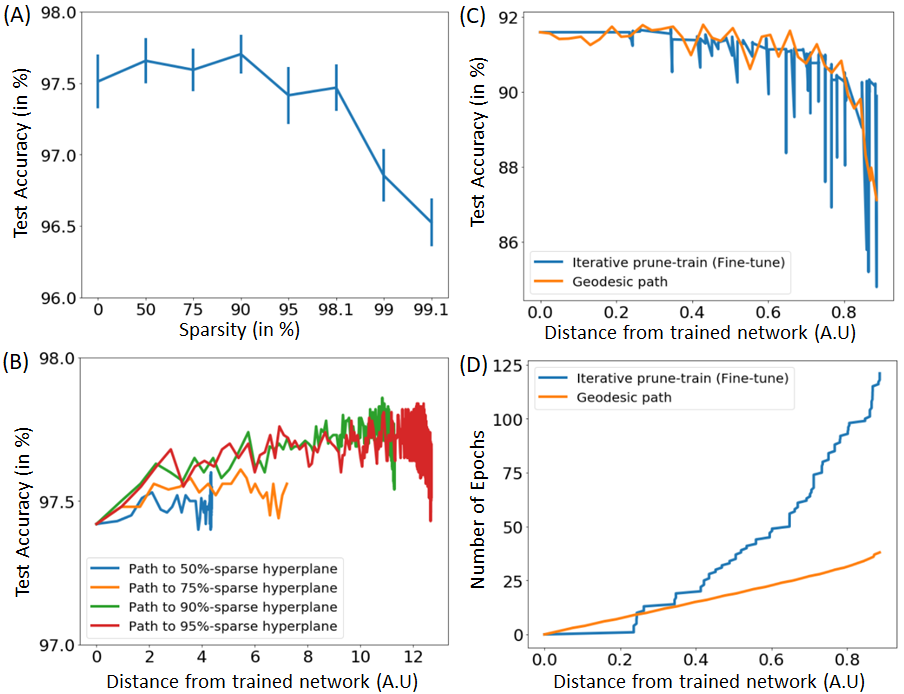}
    \caption{\textbf{Traversing geodesics to sparsify networks} (A) Test performance of sparse LeNet architectures on MNIST discovered by traversing the geodesic from the dense-MLP to the p-sparse hyperplane. The geodesic locates sparse networks that perform $\geq$97.5$\%$ for sparsities ranging from from 50 to 98.1$\%$ and finds a sparse network performing at $\sim$97$\%$ for 99$\%$ sparsity and $\sim$96.8$\%$ for 99.1$\%$ sparse network. (B) The paths traversed from the trained dense-MLP to p-sparse hyperplane (p$\in$[50,75,90,95]) are high-performing as they perform at an accuracy $\geq$97.4$\%$. (C) Test accuracy and (D) number of network update epochs for geodesic recovery (orange) vs fine-tuning (blue) of VGG11 trained on CIFAR-10, while 50 (out of 64) conv-filters are deleted from layer1 in VGG11. (D) Geodesic recovery requires $\leq$30 total update epochs, while fine-tuning requires upto 120 epochs. } 
    \label{fig:sparse_results}
    \vspace{-7mm}
\end{figure}

We demonstrate our geodesic strategy for sparsification on: (i) Multilayer perceptron (LeNet) trained on MNIST and (ii) VGG-11 trained on CIFAR-10. 
In figure-\ref{fig:sparse_results2}, we show that the geodesic strategy discovers sparse networks that perform at test accuracies comparable to the reference dense trained network $\mathbf{w_t}$. Our results (figure-\ref{fig:sparse_results}A supersede the existing benchmarks for LeNet-300-100 compression reported in literature for extreme levels of sparsity \cite{manessi2018automated, babaeizadeh2016noiseout, frankle2018lottery, alford2019training}. A comparison with the lottery ticket hypothesis \cite{frankle2018lottery} is presented in fig-\ref{fig:sparse_results2}. 

In addition to finding the p-sparse network (on $H_p$), we obtain a high-performance path connecting the dense-MLP trained on MNIST to the discovered sparse network on $H_p$. In Figure-\ref{fig:sparse_results}B, we show that 
path-connected networks from the dense network to $H_{50}$, $H_{75}$, $H_{90}$ and $H_{95}$ also perform at an accuracy $\geq$97$\%$.

Our geodesic strategy discovers structured sparse counterparts of VGG-11 trained on CIFAR-10. In Figure-\ref{fig:sparse_results}C , our geodesic approach yields high-performance paths that connect the dense-VGG11 network to its sparser counterpart wherein 50 (out of 64) conv filters from the first layer are zeroed out. We compared our strategy to traditional heuristic fine-tuning to demonstrate that our approach is both rationale and computationally efficient.  Specifically, an iterative prune-train cycle achieved through structured pruning of a single node at a time, coupled with SGD re-training requires upto 120 training epochs to identify a sparsification path. However, our geodesic strategy finds paths that quantitatively out-perform the iterative prune-train procedure and obtains these paths with only 30 training epochs (Figure-\ref{fig:sparse_results}D). 

\subsection{Alleviating Catastrophic forgetting by traversing geodesics}

Neural networks succumb to catastrophic forgetting (CF) during sequential training of tasks because training on sequential tasks alters the weights between nodes in the neural network which are locations of  "stored knowledge", resulting in the abrupt loss of  "memory" of all information from previous tasks \cite{mccloskey1989catastrophic, ratcliff1990connectionist}. Previous attempts to solve the problem of CF faced by deep networks was accomplished by meticulous tuning of network hyperparameters \cite{bengio2013empirical, srivastava2013compete} accompanied by standard regularization methods. Addressing earlier limitations, Kirkpatrick et al \cite{kirkpatrick2017overcoming} proposed elastic weight consolidation, wherein they evaluate a Fisher information matrix to guide retraining of network on a new task. 

\begin{figure}[H]
    \centering
    \includegraphics[width=1\linewidth]{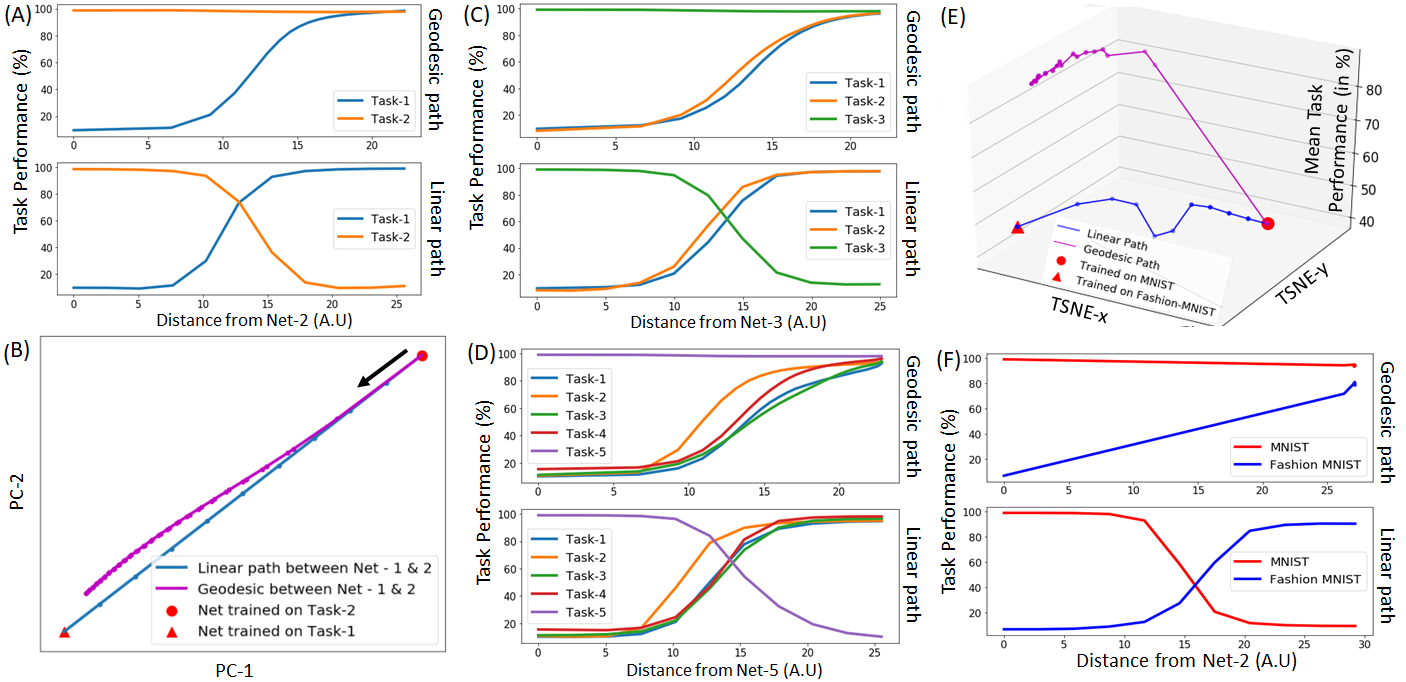}
    \caption{\textbf{Sequential Permuted-MNIST tasks: } (A) Two LeNet's are trained on permute-MNIST task-1,2 ($\mathbf{w_1}, \mathbf{w_2}$). The geodesic between $\mathbf{w_2}$ and $\mathbf{w_1}$ using metric tensor on task-2 ($\mathbf{g^2}$) discovers a network that performs at 98$\%$ test-accuracy on both tasks. The best network along the linear path performs at 80$\%$ on both tasks. (B) PCA projections of the network weights show geodesic obtained from Geo($\mathbf{w_2}$, $\mathbf{g^2}$, $\mathbf{w_1}$, $N_s$) (magenta) and linear path (blue). Both paths begin from $\mathbf{w_t} = \mathbf{w_2}$ net trained on task-2 (red circle) and move to the target $\mathbf{w_a} = \mathbf{w_1}$ net trained on task-1 (red triangle). (C) A third network pre-trained on Task-3 ($\mathbf{w_3}$) moves to target network $\mathbf{w_a}$ trained on task-1,2 obtained from (A). The geodesic finds networks that performs at [97$\%$, 97$\%$, 98$\%$] while the best network along linear path performs at [70$\%$, 60$\%$, 60$\%$] on Tasks-1,2,3 respectively. (D) A fifth network pretrained on Task-5, moves to the target network $\mathbf{w_a}$ trained on task-1,2,3,4. The geodesic path finds networks that perform at [94,95,96,96,98$\%$] on Tasks 1 to 5 respectively while the best network along the linear path performs at $\sim$60$\%$ on all 5 tasks. (E) The 'circle' and 'triangle' correspond to CNN's trained on MNIST and Fashion-MNIST datasets respectively. The x,y axes are tSNE projections of the network weights and z axes corresponds to the mean performance of the network on both, MNIST and F-MNIST. The linear path (blue) between the networks hosts networks that perform on average 40$\%$ on both tasks, while the geodesic approach discovers a curved path that performs at 94$\%$ on MNIST and 82$\%$ on F-MNIST. (F) A comparison of the linear path and geodesic approach reveals that the curved path converges at a network that simultaneously performs at 94$\%$ on MNIST and 82$\%$ on F-MNIST, while the linear path finds a network that performs at 40$\%$ on MNIST and F-MNIST.  }
    \label{fig:CF_MNISTpermute}
    \vspace{-4mm}
\end{figure}




We apply our geodesic framework in a novel fashion to mitigate CF while training networks on sequential tasks. To alleviate CF while learning $k$ sequential tasks, our goal is to discover a network that achieves a high performance on all $k$ tasks, given access to only one task at a time. Here, task is synonymous to dataset. For instance, training a network sequentially on MNIST, followed by Fashion-MNIST constitutes two sequential tasks.

We evaluate $k$ metric tensors ($\mathbf{g^1}$, $\mathbf{g^2}$, ..., $\mathbf{g^k}$) corresponding to $k$ sequential tasks (or datasets - ($X_1$,$X_2$, ..., $X_k$)). The metric, $\mathbf{g^i}$ provides a local measure of the functional distance on the pseudo-Riemannian manifold (W,$\mathbf{g^i}$), ie it measures the change in performance on task-i as the network moves on the weights space. Our geodesic approach discovers networks that perform well on sequential tasks by constructing geodesics between two locations in the weights space, one defined by network trained on the most recent task (task-i), while the other location is defined by the network trained on all previous tasks (task-1,2,..,i-1). The metric $\mathbf{g^i}$ measures the change in performance on task-i as the network moves towards a location in the weights space defined by another network trained on all previous tasks (task-1,2, ..., i-1). Therefore, our optimization strategy finds a set of path-connected networks beginning at network trained on task-i, and moving to another network trained on all previous tasks (1,2,..,i-1) with the objective of minimizing the change in performance on task-i alone. The procedure converges\footnote{One of the stopping criterion mentioned in section-2 is reached - where the network makes small oscillations towards and away from the target network.} at a network that performs well on all tasks (including most recent task-i) seen until then (task-1,2,...., i). 

\begin{wrapfigure}{r}{0.5\textwidth}
    \includegraphics[width=1\linewidth]{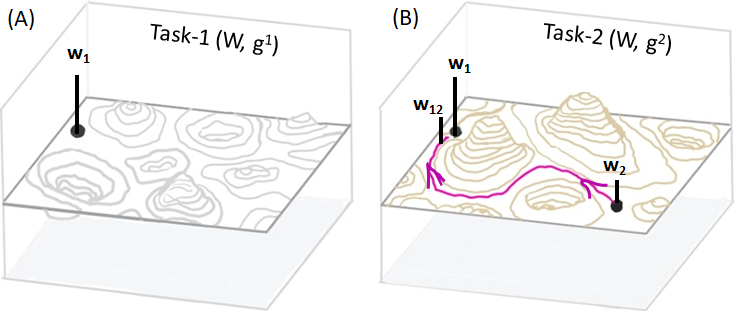}
    \caption{\textbf{Framework for mitigating catastrophic interference via geodesics} (A) Pretrained network on task-1 ($\mathbf{w_1}$) (B) Geodesic evaluated using metric on task-2 ($\mathbf{g^2}$) connecting $\mathbf{w_2}$ (pretrained on task-2) and $\mathbf{w_1}$ converges at $w_{12}$.}
    \label{fig:CF_schema}
  \vspace{-3mm}
\end{wrapfigure}
In figure-\ref{fig:CF_schema}, we illustrate our framework in action for learning two sequential tasks. We begin by training two different networks on the sequential tasks one after the other to get $\mathbf{w_1}$ and $\mathbf{w_2}$ trained on task-1,2 respectively. Subsequently, we compute the geodesic on the networks weights space, beginning from $\mathbf{w_t}$ = $\mathbf{w_2}$, network trained on task-2 by traversing the manifold (W, $\mathbf{g^2}$) towards the target network $\mathbf{w_a}$ = $\mathbf{w_1}$, network trained on task-1. The metric $\mathbf{g^2}$ measures the change in performance on task-2 as the network moves towards a location in the weights space defined by the network trained on the first task. The procedure converges at a network $\mathbf{w_{1,2}}$ that performs well on both tasks.

\vspace{-3mm}
\begin{algorithm}
\caption{Alleviating CF: Learning $k$ sequential tasks via geodesics}
\begin{algorithmic}[1]
\State Train net on Task-1 ($\mathbf{w_1}$) \Comment{Train random network on Task-1}
\State Set $i$ = 2 \Comment{Update to Task-2}
\While{$i$ $\leq$ k} \Comment{Iterate procedure for $k$ tasks}
    \State Train net on Task:i ($\mathbf{w_i}$) \Comment{Train random network on latest task}
    \State $\mathbf{w_p}$ = $\mathbf{w_{1:i-1}}$ \Comment{$\mathbf{w_p}$ := network that performs well on all previous tasks: task:(1,2,..,i-1)}
    \State $\mathbf{g} = \mathbf{g^i}$ \Comment{Compute metric tensor for most recent task (task-i)}
    \State $\gamma(t) \leftarrow Geo(\mathbf{w_i}, \mathbf{g}, \mathbf{w_p}, N_s) $ \Comment{Constructing geodesic path}
    \State $\mathbf{w_p} \leftarrow \gamma(1)$  \Comment{Terminal net of geodesic performs well on all tasks}
    \State Set $i$ = $i$ + 1 \Comment{Moving on to the next task}
\EndWhile
\end{algorithmic}
\end{algorithm}
\vspace{-3mm}

We apply our geodesic framework to the classic permuted-MNIST task \cite{le2015simple}, wherein every new task corresponds to a (fixed) unique permutation of the input pixels of the MNIST dataset. For $k$ tasks, we have $k$ permuted MNIST datasets. We use the Lenet-MLP (784-300-100-10) architecture to test our framework on sequential learning of permute-MNIST tasks.

To highlight the performance of our optimization strategy, we compare the geodesic path with the linear path between two locations in weights space: one defined by the network trained on the latest task, while the other trained on all previous tasks - obtained by iterative application of the optimization strategy. Figure-\ref{fig:CF_MNISTpermute}A shows that the network discovered along the geodesic path performs well on both tasks, and is much better than the ones uncovered by the linear path. 

We also demonstrate that our strategy can be scaled up to an arbitrary number of tasks. Fig-\ref{fig:CF_MNISTpermute} C,D, shows 5 permuted-MNIST tasks learnt sequentially without facing CF. 
Our results show improved performance over strategies like SGD (appendix) with dropout regularization and SGD with L2 regularization proposed earlier  and is comparable to EWC\footnote{Note that EWC uses a large neural network with upto 1000 nodes for their 3 task analysis, while our networks use only 400 for 5 tasks. 
}. \cite{kirkpatrick2017overcoming}. We extend our analysis to CNN's\footnote{CNN architecture used is detailed in the supplementary} trained on 2 different datasets (tasks): MNIST and Fashion-MNIST (figure-\ref{fig:CF_MNISTpermute}E,F). 


\subsection{Achieving mode connectivity via geodesics}


We apply our geodesics framework to discover high-performance paths for connecting different instances of trained DNN's (modes). As the loss landscape of DNNs are non-convex, can rely on millions of parameters and are studded with a large number of local optima and saddle points \cite{auer1996exponentially, dauphin2014identifying}, it makes the search for discovering high performance paths between the two modes of DNN's challenging. It is also observed that the linear path connecting the modes incurs high loss implying the presence of isolated local optima in the loss landscape \cite{keskar2016large, goodfellow2014qualitatively, garipov2018loss}. 


\begin{figure}[t]
    \centering
    \includegraphics[width=1\linewidth]{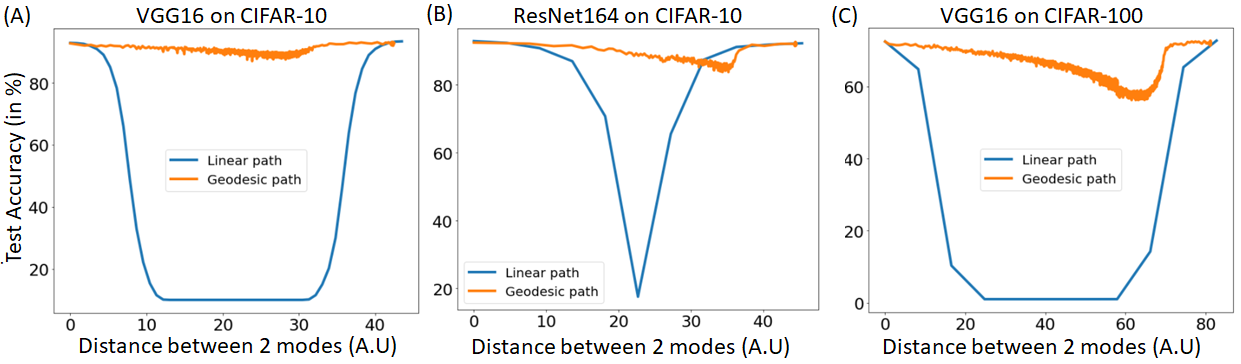}
    \caption{\textbf{Discovering high performance paths connecting network modes: } (A) The two modes of VGG16 trained on CIFAR-10 perform at 94.2$\%$ and 95.2$\%$ respectively. The linear segment (blue) connecting the modes incurs very high loss ($\sim$10$\%$) while the geodesic procedure (orange) finds a curved path with test accuracy $\geq$88$\%$. (B) The two modes of ResNet164 trained on CIFAR-10 perform at 92.44$\%$ and 92.12$\%$ respectively. The linear segment (blue) connecting the modes incurs very high loss ($\sim$20$\%$) while the geodesic procedure (orange) finds a curved path with test accuracy $\geq$85$\%$. (C) The two modes of VGG16 trained on CIFAR-100 perform at accuracy 74.4$\%$ and 75.21$\%$ respectively. The linear segment (blue) connecting the modes incurs very high loss ($\sim$1$\%$) while the geodesic procedure (orange) finds a curved path with test accuracy $\geq$55$\%$.  }
    \label{fig:modeConnect}
    \vspace{-4mm}
\end{figure}



Our experiments are performed on VGG16 trained on CIFAR-10, CIFAR-100 and 164-layer ResNet bottleneck trained on CIFAR-10. These models and datasets were chosen to demonstrate that our algorithm works on a wide range of network architectures and datasets. For each model and dataset chosen, we train two networks with different random initializations to find two mode, corresponding to two optima in the loss landscape ($\mathbf{w_1}$, $\mathbf{w_2}$). Subsequently, Geo($\mathbf{w_1}$, $\mathbf{g}$, $\mathbf{w_2}$) constructs a geodesic starting from the first mode $\mathbf{w_1}$ to the target network (second mode) $\mathbf{w_2}$, while minimizing functional difference of networks along the path on the task (CIFAR-10/100 classification), using metric tensor $\mathbf{g}$ computed on the same task. In figure-\ref{fig:modeConnect} we contrast the high-performance path obtained from the optimization strategy from the linear path that connects the two modes of (A) VGG-16 on CIFAR-10, (B) ResNet-164 on CIFAR-10 and (C) VGG-16 on CIFAR-100.

\begin{algorithm}
\caption{Discovering high performance paths connecting modes of deep networks}
\begin{algorithmic}[1]
\State Train deep network 1 ($\mathbf{w_1}$) \Comment{Training deep network with random seed 1}
\State Train deep network 2 ($\mathbf{w_2}$) \Comment{Training deep network with random seed 2}
\State $\gamma(t) \leftarrow Geo(\mathbf{w_1}, \mathbf{g}, \mathbf{w_2})$ \Comment{Geodesic from $\mathbf{w_1}$ to $\mathbf{w_2}$}
\end{algorithmic}
\end{algorithm}


\section{Discussion}

We have established a mathematical framework to construct global paths for navigating the space of neural networks to discover novel networks that have high task-performance while satisfying additional constraints on task-independent network properties. We have shown our framework in action for solving 3 major problems in ML: (i) Network sparsification, (ii) Mitigation of catastrophic interference when learning sequential tasks and (iii) Finding high-accuracy paths to connect modes of deep networks. With AI being built into many critical applications, the need for real-time processing and continuous learning on personal devices is on the rise. Our algorithm can be used for catering to this need.  In addition, we note that local processing on personal devices increases data security for the user as information remains local, without having to be streamed to the cloud. However we acknowledge that increasing accessibility of powerful hand-held AI systems in the society could have a negative impact, especially if its present in the wrong hands. Therefore, we believe widespread adoption of personal learning systems in the society should accompany educational programs for developers and subscribers regarding ethical use of AI.


\printbibliography

\section*{Checklist}

\begin{enumerate}

\item For all authors...
\begin{enumerate}
  \item Do the main claims made in the abstract and introduction accurately reflect the paper's contributions and scope?
    \answerYes{}
  \item Did you describe the limitations of your work?
    \answerYes{Mentioned in the supplementary}
  \item Did you discuss any potential negative societal impacts of your work?
    \answerYes{}
  \item Have you read the ethics review guidelines and ensured that your paper conforms to them?
    \answerYes{}
\end{enumerate}

\item If you are including theoretical results...
\begin{enumerate}
  \item Did you state the full set of assumptions of all theoretical results?
    \answerNA{}
	\item Did you include complete proofs of all theoretical results?
    \answerNA{}
\end{enumerate}

\item If you ran experiments...
\begin{enumerate}
  \item Did you include the code, data, and instructions needed to reproduce the main experimental results (either in the supplemental material or as a URL)?
    \answerYes{Code attached in supplementary}
  \item Did you specify all the training details (e.g., data splits, hyperparameters, how they were chosen)?
    \answerYes{In the supplementary}
	\item Did you report error bars (e.g., with respect to the random seed after running experiments multiple times)?
    \answerYes{wherever applicable}
	\item Did you include the total amount of compute and the type of resources used (e.g., type of GPUs, internal cluster, or cloud provider)?
    \answerYes{in the SI}
\end{enumerate}

\item If you are using existing assets (e.g., code, data, models) or curating/releasing new assets...
\begin{enumerate}
  \item If your work uses existing assets, did you cite the creators?
    \answerYes{}
  \item Did you mention the license of the assets?
    \answerNA{}
  \item Did you include any new assets either in the supplemental material or as a URL?
    \answerNA{}
  \item Did you discuss whether and how consent was obtained from people whose data you're using/curating?
    \answerNA{all datasets are traditional ML datasets(mnist, cifar, etc)}
  \item Did you discuss whether the data you are using/curating contains personally identifiable information or offensive content?
    \answerNA{}
\end{enumerate}

\item If you used crowdsourcing or conducted research with human subjects...
\begin{enumerate}
  \item Did you include the full text of instructions given to participants and screenshots, if applicable?
    \answerNA{}
  \item Did you describe any potential participant risks, with links to Institutional Review Board (IRB) approvals, if applicable?
    \answerNA{}
  \item Did you include the estimated hourly wage paid to participants and the total amount spent on participant compensation?
    \answerNA{}
\end{enumerate}

\end{enumerate}


\end{document}